\documentclass[11pt]{article}

\usepackage[margin=1in]{geometry}
\usepackage[T1]{fontenc}
\usepackage[utf8]{inputenc}
\usepackage{lmodern}
\usepackage{microtype}
\usepackage{amsmath,amssymb,amsthm,mathtools}
\usepackage{bm}
\usepackage{booktabs}
\usepackage{tabularx}
\usepackage{array}
\usepackage{multirow}
\usepackage{enumitem}
\usepackage{xcolor}
\usepackage{graphicx}
\usepackage{tikz}
\usetikzlibrary{arrows.meta,positioning,fit,calc,shapes.multipart}
\usepackage[hidelinks]{hyperref}
\usepackage[authoryear,round]{natbib}
\usepackage[nameinlink,noabbrev]{cleveref}
\usepackage{caption}
\usepackage{subcaption}
\usepackage{longtable}

\hypersetup{
  colorlinks=false,
  linkcolor=blue!60!black,
  citecolor=blue!60!black,
  urlcolor=blue!60!black
}

\title{\vspace{-1em}\textbf{Artificial Agency Program:}\\
Curiosity, compression, and communication in agents}
\author{Richard Csaky\\\texttt{richard.csaky@gmail.com}}
\date{}

\newcommand{\E}{\mathbb{E}}

\newcommand{\MI}{\mathrm{I}}
\newcommand{\DI}{\mathrm{I}^{\rightarrow}}

\begin{document}
\maketitle
\vspace{-1.2em}

\begin{abstract}
This paper presents the \emph{Artificial Agency Program} (AAP), a position and research agenda for building AI systems as reality embedded, resource-bounded agents whose development is driven by curiosity-as-learning-progress under physical and computational constraints. The central thesis is that AI is most useful when treated as part of an extended human--tool system that increases sensing, understanding, and actuation capability while reducing friction at the interface between people, tools, and environments. The agenda unifies predictive compression, intrinsic motivation, empowerment and control, interface quality (\emph{unification}), and language/self-communication as selective information bottlenecks. We formulate these ideas as a falsifiable program with explicit costs, staged experiments, and a concrete multimodal tokenized testbed in which an agent allocates limited budget among observation, action, and deliberation. The aim is to provide a conceptual and experimental framework that connects intrinsic motivation, information theory, thermodynamics, bounded rationality, and modern reasoning systems\footnote{This is a working draft. Feedback and criticism is most welcome.}.
\end{abstract}

\vspace{-0.4em}
\section{Introduction}
Contemporary frontier AI systems are increasingly capable, yet the dominant training and evaluation pipelines still under-emphasize the conditions under which biological agency develops: reality embedded interaction, finite compute and memory, constrained sensing and actuation, and the continual need to act under uncertainty \citep{SuttonBarto2018,Schmidhuber2010,OudeyerKaplanHafner2007,Friston2010,LakeEtAl2017BuildingMachines,LeCun2022PathTowardsAutonomousMI}. We start from a practical thesis: AI should be designed as part of a larger human--tool system that improves sensing, understanding, and actuation under realistic constraints, while remaining efficient and interpretable.

This framing emphasizes \emph{agency gain} and \emph{interface friction} simultaneously. Increasing raw capability alone is not sufficient if the resulting system is hard to steer, poorly coupled to human intent, or unable to operate efficiently under the time, energy, and communication budgets that dominate real environments. In many settings, the bottleneck is not whether a model can in principle provide a solution, but whether the coupled human--AI system can reliably gather information (observations), allocate limited compute, communicate intent, and act in an effective way. We therefore treat agency as a property of the coupled system, not just the standalone model \citep{ClarkChalmers1998,LiederGriffiths2020}. The strive for ever higher agency is fundamental, as it is a direct consequence of human nature defined by curiosity and playfulness aimed at discovering---predicting---the world.

A seminal work of John Archibald Wheeler postulates that physics exists due to there being someone to observe it, more precisely to ask yes/no questions \citep{itfrombit}. This motivates the curiosity objective presented in \citet{Schmidhuber2010}—increasing the rate of compression by seeking patterns (observations) on which prediction improvement is maximal given current capabilities. This avoids completely random patterns or patterns which are easy to compress and is analogous to the rate at which we can ask-and-answer questions about the universe, i.e. spatiotemporal patterns. Therefore we take this as a single core objective of human and artificial agents.

\subsection{Human frames of reference and extended agency}
Many failure modes of current systems are difficult for people to anticipate precisely because the systems are trained and deployed under non-human conditions. Internet-scale next-token training with effectively superhuman memory for textual regularities and weak grounding in action can produce competent behavior, but it does not reproduce the developmental constraints that shaped human cognition. Human intelligence is structured by limited memory, limited sensing and actuation bandwidth, limited processing speed, partial observability, and the need to rely on tools and other people for increasing agency and cognitive offloading \citep{ClarkChalmers1998,RyanDeci2000,LakeEtAl2017BuildingMachines}. Therefore parts (or dimensions) of human cognition must necessarily arise precisely due to the restrictions imposed by our specific biological instantiation from a more general intelligence manifold.

This suggests a distinction between \emph{capability} and \emph{distance from human constraints}. Two systems may achieve similar performance (intelligence) while occupying different regions of a constraint manifold defined by sensing modalities, action affordances, energetic pressure, temporal scale, and memory budget. We propose measuring both overall competence (e.g., predictive performance, empowerment, control) and constraint proximity, because these jointly affect interpretability, collaboration quality, and failure modes \citep{Chollet2019,LiederGriffiths2020}. By representing intelligence as such a manifold, one may quantify overall competence as the magnitude of a point on the manifold, providing a scalar comparison between human and artificial intelligence. Human-likeness (or alienness) can be measured with a distance metric between human and AI points on the manifold.

The reason it is hard to encode into AI the fundamental aspects of human intelligence is that they are part of a fundamentally human frame of reference. What is curious or novel to us is defined by our scale in time and space, by evolutionary programming, by biological considerations, by the properties of our sensing and actuating mechanisms. Indeed even by our level of intelligence or cognition---testament to this is the fact that looking at the arch of human (and animal) history, curiosity and inquisitiveness has always shifted—adjusted—depending on our cognitive abilities. A human 1000 years ago could hardly be curious about differential topology. An agent can only be curious with regards to what it can sense, understand and act upon—where all three include tools forming part of the extended self. It is erroneous to simply jump-start intelligence without considering the manifold of constraints it is embedded in.

\subsection{Curiosity, compression, and progressive capability expansion}
We adopt a learning-progress view of curiosity: intrinsically valuable patterns are neither random nor trivial, but those for which the agent can currently improve compression or prediction \citep{Schmidhuber2010,OudeyerKaplanHafner2007}. This makes curiosity inherently capability-relative. An agent can only be curious about patterns it can (in some sense) sense, model, or act upon, and the difficulty of useful tasks should grow with the agent's capabilities rather than being fixed in advance. This view aligns with developmental perspectives in which exploration and play are active processes of skill acquisition rather than passive consumption of novelty \citep{RyanDeci2000,OudeyerKaplanHafner2007}.

On this view, curiosity creates pressure for \emph{capability expansion}. If learning progress is limited by sensing, actuation, memory, or compute, the agent should prefer interventions that relax whichever bottleneck most improves future learning progress, subject to cost. This links curiosity to empowerment, control, and communication: an agent may benefit from seeing more, doing more, delegating more, or coordinating with others and tools more effectively. Because the world presents patterns over space and time, this pressure naturally favors longer-horizon prediction and increasingly structured internal models, not merely myopic novelty seeking \citep{Schmidhuber2010,HaSchmidhuber2018WorldModels,HafnerEtAl2023DreamerV3}. Due to this pressure of predicting as far as possible into the future, the study of mathematics, physics, information, etc., should naturally emerge from such intrinsically motivated agents, as they aim to compress spatiotemporal observations into as small a description as possible. The environment is crucial: to develop truly capable agents we need them to be embedded in the physical spatiotemporal reality, not in a computer sandbox. The learning trajectory and actions at any step are constrained by the physical sensing, actuating, and energy resources of the agent.

For example, an agent living in a 2D world may be "frustrated" (and therefore curious) that it cannot “see” a complete circle, only parts of it. Similarly as a human I may be annoyed that I cannot see 3D objects in their entirety at the same time, whereas I can see 2D shapes wholly. The implications are: 1. Since I am currently seeing the world in 3D I may have a desire to see 4D objects, but my desire to see 10D objects is much lower, and 2. I cannot predict the kinds of goals or desires that seeing in 4D (i.e., gaining more sensing capability) would unlock. A similar analogy can be made with respect to actuation and cognition. Perhaps an agent that is aware of its limitations will be most curious about what happens when removing some of those limitations, since this is a future fundamentally harder to predict than asking questions and answering within its current limitations. It is a sort of meta-objective: I not only want to predict the world as I can interact with it now, but I want to predict the kinds of questions I may ask when my capabilities increase. Analogously, my true underlying objective is to predict the hidden (unobservable) state of the environment in its totality.

\subsection{Energy--performance frontiers}
Efficiency is a natural concern in AAP. Many AI systems become practically useful only when they solve tasks within meaningful time and energy budgets. The guiding intuition is closer to engineering flight than to copying bird biomechanics: progress comes from discovering designs that solve the task under constraints. This motivates explicit energy/compute--performance frontiers and MDL-style criteria for how much task performance is obtained per unit cost \citep{Rissanen1978,Grunwald2007,Hutter2005,PattersonEtAl2021Carbon}.

We therefore emphasize not only peak performance, but adaptive allocation of computation and communication. In many tasks, the optimal policy is not to spend maximal compute at every step, but to modulate effort based on uncertainty, expected value of information, and opportunity cost \citep{Graves2016ACT,BaninoBalaguerBlundell2021PonderNet,CallawayEtAl2018UAI,LiederGriffiths2020}. This efficiency emphasis also motivates the later focus on private deliberation tokens, action tokens, and observation tokens as interchangeable budgeted resources.

\section{Formal Setup}
We model an embedded agent interacting with an environment as a partially observed controlled process. Let \(X_t\) denote the (generally hidden) environment state, \(S_t\) the agent internal state (memory + policy state ($\theta_t$) + latent world-model state), \(O_t\) the observation, and \(A_t := (U_t, V_t)\) the action.

\begin{align}
O_t &\sim p_O(\cdot \mid X_t;\, c^O_t), \\
S_{t} &\sim p_S(\cdot \mid S_{t-1}, O_t, U_{t-1}, V_{t-1};\, c^C_t), \\
(U_t, V_t) &\sim \pi(\cdot \mid S_t;\, c^A_t), \\
X_{t+1} &\sim p_X(\cdot \mid X_t, U_t),
\end{align}
where \(c^O_t, c^A_t, c^C_t\) denote observation, actuation, and compute/interface capacity constraints (e.g., sensor bandwidth, action alphabet/determinism, context length, update budget). These capacities themselves may be modeled as dynamic and agent-controllable.

\begin{align}
c^O_{t+1} &\sim p_{c^O}(\cdot \mid c^O_{t}, V_t), \\
c^A_{t+1} &\sim p_{c^A}(\cdot \mid c^A_{t}, V_t), \\
c^C_{t+1} &\sim p_{c^C}(\cdot \mid c^C_{t}, V_t).
\end{align}

We view $p_S$ as the agent-side update operator and allow it to include both recurrent/memory updates and (optionally) learning updates of parameters $\theta_t$ (e.g., a bounded number of gradient/Bayes-style update steps), all priced under the compute/interface budget $c^C$.

\paragraph{Learning-progress intrinsic reward.}
We adopt a Schmidhuber-style learning-progress signal, not raw novelty \citep{Schmidhuber2010}. Let \(\mathcal{L}_{\text{pred}}\) be a predictive loss for future observations over horizon \(H\). For any integers $a<b$, define the interaction segment
\begin{equation}
\mathcal{D}_{a:b}
\;:=\;
\big( O_a, A_a, O_{a+1}, A_{a+1}, \ldots, O_{b-1}, A_{b-1}, O_b \big),
\label{eq:segment_def}
\end{equation}
where $A_t=(U_t,V_t)$ as in the formal setup.

Let $p_\theta$ denote the agent's predictive model for observations, which maps a history to a distribution over the next observation:
\begin{equation}
p_\theta\!\left(O_{t+1}\mid \mathcal{D}_{a:t}\right)
\;\equiv\;
p_\theta\!\left(O_{t+1}\mid O_{a:t}, A_{a:t}\right).
\label{eq:pred_model_def}
\end{equation}
We instantiate the horizon-$H$ predictive loss as the cumulative negative log-likelihood
\begin{equation}
\mathcal{L}_{\text{pred}}(\theta;\mathcal{D}_{t:t+H})
\;:=\;
\sum_{h=1}^{H}
\ell\!\left(
p_\theta(\cdot \mid O_{t:t+h-1}, A_{t:t+h-1}),
\, O_{t+h}
\right),
\qquad
\ell(p,o):=-\log p(o).
\label{eq:Lpred_def}
\end{equation}

A generic intrinsic reward is then
\begin{equation}
r_t
\;:=\;
\mathcal{L}_{\text{pred}}(\theta_{t-H-1};\mathcal{D}_{t-H:t})
\;-\;
\mathcal{L}_{\text{pred}}(\theta_{t-H};\mathcal{D}_{t-H:t}),
\qquad t>H,
\label{eq:learning_progress}
\end{equation}
where $r_t := 0 \text{ for } t\le H$.
This rewards \emph{improvement} in predictive compression rather than complexity or surprise alone, consistent with learning-progress formulations and modern world-model practice \citep{Schmidhuber2010,HaSchmidhuber2018WorldModels,HafnerEtAl2023DreamerV3}.

We combine intrinsic reward with costs:
\begin{equation}
J(\pi, p_S) \;=\; \E\!\left[\sum_{t=1}^{T}
\gamma^{t-1}
\Big(r_t
-\lambda_O C_O(t)
-\lambda_E C_E(t)
-\lambda_C C_C(t)
-\lambda_M C_M(t)
\Big)\right].
\label{eq:overall_objective}
\end{equation}
Where:
\begin{itemize}
\item \(C_O(t)\): observation-processing cost (e.g., number of observation tokens/bits ingested or sensor bandwidth used at time 
$t$),
\item \(C_E(t)\): actuation/maintenance cost (e.g., environment-defined energy penalty for executing $U_t$),
\item \(C_C(t)\): compute/deliberation/learning cost (e.g., FLOPs, number of deliberation tokens, or number of update steps executed at time $t$)
\item \(C_M(t)\) a memory-maintenance cost (motivated by finite-state maintenance and thermodynamic viewpoints on information processing \citep{StillSivakBellCrooks2012,ParrPezzuloFriston2022}).
\end{itemize}

\subsection{Control, empowerment, and plasticity}
AAP’s hypotheses relate predictive compression to control and empowerment.

\paragraph{Empowerment (actuation capacity into future observations).}
A standard \(k\)-step empowerment definition is the channel capacity from action sequences to future observations \citep{KlyubinPolaniNehaniv2005,KlyubinPolaniNehaniv2005Ecal,SalgeGlackinPolani2014}:
\begin{equation}
\mathcal{E}_k(s_t)
\;=\;
\max_{p(a_{t:t+k-1})}
\MI\!\big(A_{t:t+k-1}; O_{t+k} \,\big|\, S_t=s_t\big).
\label{eq:empowerment}
\end{equation}

\paragraph{Plasticity (observation-to-action influence).}
We emphasize ``plasticity'' as environment pressure on the agent. A natural proxy is directed information from observations to actions \citep{Massey1990,Kramer1998, abel2025plasticity}:
\begin{equation}
\mathcal{P}_{t:T}
\;:=\;
\DI(O_{t:T} \to A_{t:T}) \;=\; \sum_{\tau=t}^{T} \MI(O_{t:\tau}; A_\tau \mid A_{t:\tau-1}).
\label{eq:plasticity}
\end{equation}
High $\mathcal{P}_{t:T}$ can reflect beneficial adaptivity/reactivity. To interpret it specifically as \emph{costly learning pressure} (as opposed to purely reactive control), we report it alongside an explicit update proxy such as $\|\theta_t-\theta_{t-1}\|$ or the number of learning/update steps executed under the compute budget.

\paragraph{Predictive sufficiency / compression.}
We distinguish:
(i) predictive performance on future observations,
(ii) compression of hidden state \(X_t\) (typically only approximable),
and (iii) model complexity / description length of the internal state or policy \citep{Rissanen1978,Grunwald2007,TishbyPereiraBialek1999,CoverThomas2006,BengioCourvilleVincent2013Representation}. This avoids collapsing all notions of ``understanding'' into a single score.

\subsection{Unification as interface quality}
We introduce \emph{unification} as reduction of sensing/acting bottlenecks between agent and environment. We formalize unification as an interface-quality concept.

Let \(b_O,b_A\) denote observation and action bottleneck parameters (coarsening, noise, latency, cardinality constraints, etc.). These bottleneck parameters are simply an alternative parameterization of the capacity constraints defined earlier: $b_O$ indexes the observation channel family $p_O(\cdot\mid X; b_O)$ with effective capacity $c^O(b_O)$, and $b_A$ indexes the actuation/policy interface with effective capacity $c^A(b_A)$. We likewise let $b_L$ parameterize a communication/self-communication interface whose cost is accounted for under the same budgeting terms already present in \cref{eq:overall_objective}. Define a task-relative \emph{unification score}
\begin{align}
\mathcal{U}_t
\;:=\;
w_O\, \underbrace{\Big(1-\frac{\mathcal{R}_O(b_O)}{\mathcal{R}_O^{\max}}\Big)}_{\text{observation losslessness proxy}}
&+
w_A\, \underbrace{\Big(1-\frac{\mathcal{R}_A(b_A)}{\mathcal{R}_A^{\max}}\Big)}_{\text{action authority proxy}}
+
w_L\, \underbrace{\Big(1-\frac{\mathcal{R}_L(b_L)}{\mathcal{R}_L^{\max}}\Big)}_{\text{communication bottleneck proxy}},\\
w_O + w_A + w_L &= 1,\qquad  \mathcal{R}_\cdot\ \geq 0 
\label{eq:unification_proxy}
\end{align}

where \(\mathcal{R}_\cdot\) are calibrated inefficiency measures induced by each bottleneck. The exact choice of \(\mathcal{R}\) is environment-specific. This way we can test whether improved agents systematically spend resources to increase effective interface quality when allowed to do so.

One way to define the observation-interface inefficiency is equivocation:
\begin{equation}
\mathcal{R}_O
\;:=\;
H(X|O), \qquad \mathcal{R}^{\max}_O = H(X).
\label{eq:RO_KL}
\end{equation}

$\mathcal{R}_A$ and $\mathcal{R}_L$ can be similarly defined.

\section{Hypotheses}

\subsection{H1: Pragmatic alignment of objectives}
In resource-bounded embedded agents, interventions that increase learning progress on future observations also tend to increase useful control over task-relevant environmental degrees of freedom, and vice versa, over a substantial regime of tasks and constraints. Moreover, optimizing \cref{eq:overall_objective} with respect to agent-side observations implies similar learning progress (or control) over the true environment hidden state from which observations are drawn, when interfaces and bottlenecks are flexible enough to allow for losslessness (i.e. invertibility) in the limit.

This is weaker than strict equivalence between predictive compression, hidden-state compression, and control. It predicts alignment over a regime, not identity for all environments. It is compatible with counterexamples where prediction is easy but uncontrollable, or control is high but not informative.

\subsection{H2: Boundary pressure toward unification}
When an agent can invest resources to modify its sensing/acting/communication interfaces, optimization of \cref{eq:overall_objective} will allocate resources toward interface improvements that increase long-horizon learning progress and control, yielding monotonic gains in task-relative unification \(\mathcal{U}\) until costs dominate. In finite environments and with favorable scaling of interface cost, a stronger boundary-collapse regime may emerge asymptotically. In an idealized limit where (i) environment complexity is finite and (ii) marginal interface-improvement costs scale favorably, the optimizer can drive $\mathcal{U}\to 1$ (maximal effective interface quality). This as a limit-case conjecture about the objective’s asymptotics, not a practical prediction for real systems.

\subsection{H3: Constraint-induced predictive/control pressure}
Under continued viability constraints (cannot stop interacting), coarse/noisy interfaces, and nonzero costs for action, memory maintenance, and frequent policy updates, agents are driven toward better prediction and selective control because these reduce costly plasticity and wasted computation. Therefore even without directly providing the intrinsic reward of \cref{eq:learning_progress}, an agent may nevertheless be forced to optimize toward it in realistic environments.

This hypothesis aligns naturally with information-theoretic and thermodynamic arguments linking predictive state representations to energetic efficiency \citep{StillSivakBellCrooks2012,Friston2010,ParrPezzuloFriston2022,OrtegaBraun2013ThermodynamicsDecision}.

\subsection{H4: Adaptive compute optimality}
We further predict that agent performance under fixed budgets depends not only on model capacity, but on how computation is allocated over time. A meta-controller that dynamically allocates observation, action, and deliberation compute should outperform fixed schedules under the same total budget, especially when task difficulty and observability vary across timesteps. This connects to adaptive computation and metareasoning \citep{Graves2016ACT,BaninoBalaguerBlundell2021PonderNet,CallawayEtAl2018UAI,LiederGriffiths2020}.

\subsection{H5: Self-communication bottleneck}

An explicit self-communication channel (e.g., text-like or symbol-like private tokens) can improve performance and sample efficiency on tasks requiring long-horizon credit assignment, compositional planning, or coordination, beyond what latent recurrence/planning alone achieves, provided the private channel is bandwidth-regularized. Importantly, the claim is not that verbalized traces are always optimal, but that \emph{selective} self-communication may be an efficient option in a broader action space that also includes direct action and latent deliberation \citep{YaoEtAl2023TreeOfThoughts,KimKimThorne2025Pause,XieEtAl2025MiniOmniReasoner,WangPengLiu2026PLAT}.

\begin{table}[t]
\centering
\caption{Hypotheses with positive predictions and disconfirmation paths.}
\label{tab:hypotheses}
\small
\begin{tabularx}{\textwidth}{@{}p{1.1cm}p{7.0cm}X@{}}
\toprule
\textbf{Hyp.} & \textbf{Positive evidence} & \textbf{Disconfirmation / boundary case} \\
\midrule
H1 & Across tasks, interventions improving learning progress also improve control/empowerment metrics under fixed budgets. & Predictive gains with no corresponding control gain on key tasks; or control gains via exploitation that degrade predictive competence. \\
H2 & Agent spends budget to widen sensors/actions/communication only when long-horizon return improves; \(\mathcal{U}\) rises then saturates. & Agent never invests in interface improvements despite clear gains; or always over-invests regardless of cost. \\
H3 & Under stronger cost/viability constraints, policies shift toward better predictive state and reduced costly reactive plasticity. & Constraint increases merely collapse behavior without improved predictive organization or selective control. \\
H4 & Meta-control beats fixed observe/act/deliberate schedules at equal budget. & No gain over tuned static schedules, implying control overhead dominates. \\
H5 & Private self-communication tokens improve task performance / sample efficienc. & Private channel collapses into redundant verbosity, or latent-only recurrence dominates under equal compute. \\
\bottomrule
\end{tabularx}
\end{table}

\section{Language as an Information Bottleneck}
We view language as a selective, lossy, resource-constrained communication channel whose value depends on the environment and on the presence of other agents/tools.

This perspective is consistent with extended cognition \citep{ClarkChalmers1998}, communication under resource rationality \citep{LiederGriffiths2020}, and recent reasoning systems that separate or modulate explicit verbal traces from latent computation \citep{Graves2016ACT,BaninoBalaguerBlundell2021PonderNet,WangPengLiu2026PLAT}. It also helps separate two questions that are often conflated in language-model practice: whether language is useful as a training substrate for acquiring broad abstractions, and whether language is the best \emph{online} medium for every intermediate computation once an agent is acting in a world.

From the perspective of a single embedded agent, language is one possible compression channel among many. Its value increases sharply when the environment contains other agents, humans, and tools whose behavior can be altered through communication. In such settings, communication is simply another action class whose utility depends on whether the expected future gain in prediction, control, or coordination exceeds its cost in time, bandwidth, and energy. We therefore treat communication as an instrumentally selected action, not a universally privileged mode.

We also takes seriously the possibility that language-like communication can be useful \emph{internally}. A compact, lossy self-directed channel may support deliberation by forcing the system to encode intermediate structure in a form that is easier to re-read, monitor, and compose. One plausible mechanism is a communicator--interpreter loop: to produce useful messages (even to itself), the agent must model how those messages will be parsed, which can induce a partial third-person stance on its own intermediate states. This motivates concrete experiments comparing latent-only recurrence with explicit private token channels (i.e. standard reasoning traces) under matched cost.

\subsection{Implications for LLMs}
All LLM-style (including VLMs and VLAs) models should have reasoning modules after having "learned" language. The purpose of a reasoning module however is not to produce a stream-of-words, but rather to allow for deliberation without action. Therefore when a model is post-trained for reasoning on various tasks (including language itself) it should be able to produce sequential hidden state updates without the use of language tokens. Such a model would have the flexibility to choose when to produce and when to avoid self-communication through language and do pure deliberation (without tokenization bottlenecks) instead. Additionally, deliberation can be expanded to any token modality, including vision and action tokens. This is analogous to visual or motor imagery in neuroscience.

Such a model might observe various types of input tokens and be able to produce and interleave these at any step. Importantly during production there is no fixed order, similar to how a human might pause mid-sentence to do a tiny deliberation loop (with or without explicit language-thoughts). Allowing the model to flexibly start “thinking” while it has not finished a paragraph, or continue thinking while producing an output could help with error recovery and self-monitoring.

In the most general case we propose a modality-agnostic token taxonomy with three roles per modality \(m\in\{V,S,T,A,\ldots\}\):

\[
m_i \text{ (input)}, \qquad m_s \text{ (private/self)}, \qquad m_o \text{ (public/output)}.
\]

This yields a unified interface where the policy can interleave incoming observations, external actions, and internal deliberation:
\[
\underbrace{V_i,S_i,T_i}_{\text{incoming}} \;\;\Leftrightarrow\;\;
\underbrace{H, V_s,S_s,T_s,A_s}_{\text{deliberation / imagery / reasoning}}
\;\;\Leftrightarrow\;\;
\underbrace{V_o,S_o,T_o,A_o}_{\text{communication / output}},
\]

where V denotes vision, S denotes audio, T denotes text, and A denotes action tokens. H refers to the output hidden state. The key design question is not whether to allow private tokens, but \emph{when they are worth emitting} under budget, and \emph{which modality is most effective} for deliberation at any given step. This directly supports our compute-efficiency criterion.

It is an open and difficult question how to train such a flexible model given current LLM paradigms. One way to train it for hidden-state deliberation would be to choose blanks (i.e. reserved non-language tokens) whenever the confidence in the next text token is below a threshold. For example one can take a reasoning trace written by a human and run it through a base LLM to get confidence scores for each token. Any position whose score is below a threshold gets shifted to the right and a "blank" is inserted. Then the LLM can be fine-tuned on these augmented traces for a while, and subsequently use the fine-tuned model to regenerate scores for our augmented traces, inserting new blanks according to confidence scores, and so on. After each step the number of positions where blanks are included should decrease and perhaps there is an optimum where no more blanks are required. We should expect the final trace to have many blanks in the beginning and at conceptual jumps in the reasoning. This procedure is still somewhat artificial and not necessarily in line with the goal of using blanks for truly free deliberation. Ultimately, choosing between blanks and other types of imagery modalities (including text) should not be based on language modeling score.

\section{Experimental Agenda}
AAP is intended to be tested in stages, starting with environments where observability, controllability, and cost structure are explicit, and then moving toward richer interactive multimodal settings. The objective is not only to maximize a given benchmark score, but to validate the proposed metrics, identify counterexamples, and characterize the regimes in which the hypotheses hold.

\subsection{Stage 1: Synthetic POMDPs}
The first stage should use toy but diagnostic partially observed environments with known latent dynamics and controllable bottlenecks. Gridworld-like domains with sensor coarsening, latency, observation noise, and switchable action-set cardinality are particularly useful because they permit direct intervention on the variables that define AAP's unification and cost definitions. This stage is primarily for calibrating the metrics, stress-testing H1--H3, and identifying degenerate regimes (e.g., environments where prediction is cheap but irrelevant, or control is high but uninformative).

\subsection{Stage 2: ARC-AGI-style interactive inference}
ARC-AGI-3-style tasks are attractive because they emphasize compositional priors and generalization under sparse data \citep{Chollet2019}. Here we can directly test interactive perception--action--deliberation with explicit compute and self-communication costs. This shift makes it possible to study when additional observation is worth requesting, when internal computation is worth spending, and when direct action is preferable. It also creates a clean setting for evaluating self-communication channels.

\subsection{Stage 3: Multimodal VLA meta-control}
The third stage introduces a pretrained multimodal model as a perception and world-model backbone, together with a lightweight meta-controller that decides at each step whether to acquire more input, act on the environment, or spend budget on private deliberation and adaptation. A frozen backbone plus lightweight adaptation (e.g., LoRA, recurrent adapters, or external memory) is a practical starting point; richer online learning can be introduced later as compute allows. This stage makes the AAP hypotheses concrete in a setting that is close to current multimodal systems.

\subsection{Efficiency and pareto optimality}
A useful diagnostic is an energy/compute--performance frontier. Let \(P\) be task performance and \(C\) a normalized cost. We can summarize each policy by \((C,P)\) and compare to an empirical frontier \(\mathcal{F}\). One scalar score is L2 distance to the frontier:
\begin{equation}
d_{\mathcal{F}}(C,P) \;=\; \inf_{(c,p)\in\mathcal{F}} \big\| (C,P)-(c,p)\big\|_2,
\end{equation}
or a task-weighted regret relative to frontier points at matched cost. This formalizes one of our core intuitions: intelligent systems should allocate compute adaptively, not just maximize score irrespective of expenditure.

Importantly in the case of large, pre-trained models the cost should include not just inference but the total pre-training energy usage. Ideally AI systems should be able to dynamically adjust energy usage in a pareto-optimal way, i.e. a particular system's distance from the pareto-optimal curve is the same at any point. A SOTA system then should provide an upper envelope on all previous models across varying levels of energy usage.

\subsection{Proof of concept study}
A minimal first study can be executed with modest resources by using an ARC-AGI-3-like grid environment with a hidden state and patchwise observations. The agent interacts through three meta-actions---\texttt{observe}, \texttt{act}, and \texttt{deliberate}---and each emitted or consumed token incurs an explicit cost. Observation and action channels can be bottlenecked by cardinality limits, noise, cropping, or latency, allowing direct manipulation of the quantities used in the unification proxy.

For the model backbone, a compact transformer with image-token and text-token support is sufficient in the first instance. The meta-controller can be a lightweight sequence model over the three meta-actions, trained with a policy-gradient or bandit-style objective on \cref{eq:overall_objective}. A practical initialization is to keep the backbone frozen, treat private deliberation as either latent recurrence or private tokens, and permit sparse low-rank adaptation only after the meta-controller behavior stabilizes. Additional actions aimed at modifying interface bottlenecks can be included.

The key comparisons are straightforward: adaptive meta-control versus fixed observe/act/deliberate schedules; latent-only recurrence versus explicit private tokens; and tight versus loose observation/action bottlenecks, each with and without deliberation cost penalties. Success should not be defined solely by raw task score. The primary criteria are frontier improvement at matched cost, nontrivial and interpretable use of private computation, and systematic shifts in meta-action allocation as interface bottlenecks and costs change. These outcomes would provide direct evidence about H4--H5 while also producing diagnostic signals for H1--H3.

\section{Related Work}
AAP is most directly aligned with the intrinsic-motivation and developmental-robotics literature, especially work that treats curiosity as \emph{learning progress} rather than raw novelty or surprise \citep{Schmidhuber2010,OudeyerKaplanHafner2007}. This distinction is central to the program: novelty-only signals invite stochasticity seeking, whereas learning-progress signals favor patterns that are currently learnable but not yet mastered. Modern exploration methods based on prediction error or novelty remain highly relevant as practical baselines and components, but AAP adopts learning-progress style objectives because the agenda is explicitly concerned with the rate of improvement in predictive compression, not merely exposure to unexpected inputs \citep{PathakEtAl2017,BurdaEtAl2019,Schmidhuber2010}.

A second pillar is empowerment and information-theoretic agency. Empowerment formalizes agent-centric control as a channel-capacity-like quantity from action sequences to future observations and provides a principled bridge between control, exploration, and interface design \citep{KlyubinPolaniNehaniv2005,KlyubinPolaniNehaniv2005Ecal,SalgeGlackinPolani2014,JungPolaniStone2011}. AAP extends this perspective by explicitly pairing empowerment with a complementary observation-to-action quantity (plasticity, proxied here by directed information) and by embedding both within a resource-cost objective. This pairing is useful because the same behavior can look desirable or undesirable depending on whether frequent reactive updates are cheap or expensive \citep{Massey1990,Kramer1998}.

AAP also draws heavily on predictive-processing, active-inference, and world-model traditions, all of which emphasize predictive internal structure and action under uncertainty \citep{RaoBallard1999,Friston2010,AitchisonLengyel2017WithOrWithoutYou,ParrPezzuloFriston2022,HaSchmidhuber2018WorldModels,HafnerEtAl2023DreamerV3}. Most relevant is the idea that good internal models support efficient action and adaptation. AAP differs mainly in emphasis: it foregrounds explicit sensing, acting, communication, and compute bottlenecks as manipulable interface variables, and it studies how agents allocate resources across these channels rather than focusing only on the representational objective.

The agenda further overlaps with information-theoretic and thermodynamic treatments of learning and decision-making. The thermodynamics-of-prediction results of Still and colleagues motivate the coupling between predictive information and dissipative cost \citep{Still2009,StillSivakBellCrooks2012}, while related work on thermodynamic metrics and bounded-rational decision-making clarifies how information-processing costs can shape optimal behavior \citep{SivakCrooks2012,OrtegaBraun2013ThermodynamicsDecision}. AAP imports this intuition at the level of design pressure: under viability constraints, partial observability and control, and nonzero costs for memory maintenance, action, and adaptation, prediction and selective control become economically important, not merely descriptively useful.

On the representation side, AAP is consistent with MDL and information-bottleneck perspectives, which treat useful representations as compressed summaries that preserve task- or prediction-relevant structure \citep{Rissanen1978,Grunwald2007,TishbyPereiraBialek1999,CoverThomas2006,Hutter2005,BengioCourvilleVincent2013Representation}. Our contribution is a synthesis: predictive compression is evaluated together with control, empowerment, and interface quality under explicit budgets. This matters because predictive compression alone does not determine whether a system can alter the variables it predicts, and control alone does not ensure that the controlled variables are informative or broadly useful.

Our discussion of language and self-communication intersects several strands of work: extended cognition and cognitive offloading \citep{ClarkChalmers1998}; resource-rational cognition and metareasoning \citep{LiederGriffiths2020,CallawayEtAl2018UAI}; adaptive computation mechanisms that decide when to spend additional internal computation \citep{Graves2016ACT,BaninoBalaguerBlundell2021PonderNet}; and recent reasoning systems that partially decouple internal deliberation from public verbalization \citep{YaoEtAl2023TreeOfThoughts,KimKimThorne2025Pause,XieEtAl2025MiniOmniReasoner,WangPengLiu2026PLAT}. Our position is perhaps narrower and more operational: language-like traces are one optional channel in a broader token/action budget, and their value should be measured relative to latent alternatives under matched compute and communication cost.

Finally, AAP is related to benchmark and systems-level discussions of intelligence that emphasize priors, generalization, and real-world constraints. ARC-AGI-style framing highlights the importance of priors and skill-acquisition efficiency \citep{Chollet2019}; broader arguments such as ``Reward is Enough'' emphasize the generality of reward-driven optimization mechanisms \citep{SilverEtAl2021RewardIsEnough}; and discussions of autonomous machine intelligence and human-like learning emphasize the need for richer world models and action-grounded training regimes \citep{LakeEtAl2017BuildingMachines,LeCun2022PathTowardsAutonomousMI}.

\section{Limitations}
We highlight several structural limitations and invite the community for feedback and suggestions for improvement. First, non-equivalence between prediction, empowerment, and task reward is common rather than exceptional; the agenda explicitly expects regimes in which these quantities diverge because of reward misspecification, hidden confounders, or severe partial observability. Second, some of the most interesting quantities in AAP---especially empowerment and directed information in high-dimensional settings---are expensive to estimate and may require loose variational bounds, which can blur interpretation.

A third limitation concerns measurement of energy and efficiency. FLOPs, wall-clock time, and token count are convenient but imperfect proxies for physical energy expenditure, and cross-system comparisons are sensitive to hardware, implementation details, and parallelism \citep{PattersonEtAl2021Carbon}. A fourth limitation is that self-communication channels can become brittle or degenerate into verbose private codes unless regularized by explicit bandwidth costs, latency constraints, or downstream utility. Finally, human-likeness remains intrinsically multi-objective: similarity in constraints may improve interpretability in some settings, but it does not automatically imply safety or desirable behavior across tasks. Our manifold view of intelligence highlights the need for better fundamental understanding and science behind how specific human-related constraints and frames of reference shape and define human intelligence and how the broader manifold looks like.

\section{Conclusion}
We make the argument for treating AI systems as \emph{embedded, resource-bounded agents} whose usefulness is best understood at the level of the coupled human--tool--environment system. We formalized this stance with (i) curiosity-as-learning-progress as an intrinsic development signal, (ii) explicit budgets over observation, actuation, compute, and memory in a single objective (\cref{eq:overall_objective}), (iii) complementary proxies for prediction, empowerment/plasticity, and (iv) \emph{unification} as an experimentally manipulable notion of interface quality (\cref{eq:unification_proxy}), including language and self-communication as selective information bottlenecks rather than universally privileged computation.

AAP is a call to make the tradeoffs that already govern real deployment \emph{explicit and measurable}: what to observe, when to think, what to communicate, and how much energy and latency are worth spending for marginal gains in prediction and control. We hope this framing helps shift discussion from single-score capability to a more grounded science of agency under constraints, and we invite criticism, counterexamples, and sharper theoretical framing and experimental designs.

\bibliographystyle{plainnat}
\bibliography{aap}

\end{document}